# Performance Analysis of Enhanced Clustering Algorithm for Gene Expression Data

T.Chandrasekhar[1], K.Thangavel[2] and E.Elayaraja[3]

[1] Research Scholar, Bharathiar university,
Coimbatore, Tamilnadu, India - 641 046.

[2] Department of Computer Science, Periyar University,
Salem, Tamilnadu, India -636 011.

[3] Department of Computer Science, Periyar University,
Salem, Tamilnadu, India -636 011.

**Abstract**
Microarrays are made it possible to simultaneously monitor the expression profiles of thousands of genes under various experimental conditions. It is used to identify the co-expressed genes in specific cells or tissues that are actively used to make proteins. This method is used to analysis the gene expression, an important task in bioinformatics research. Cluster analysis of gene expression data has proved to be a useful tool for identifying co-expressed genes, biologically relevant groupings of genes and samples. In this paper we applied K-Means with Automatic Generations of Merge Factor for ISODATA- AGMFI. Though AGMFI has been applied for clustering of Gene Expression Data, this proposed Enhanced Automatic Generations of Merge Factor for ISODATA- EAGMFI Algorithms overcome the drawbacks of AGMFI in terms of specifying the optimal number of clusters and initialization of good cluster centroids. Experimental results on Gene Expression Data show that the proposed EAGMFI algorithms could identify compact clusters with perform well in terms of the Silhouette Coefficients cluster measure.

## 1. Introduction

Clustering has been used in a number of applications such as engineering, biology, medicine and data mining. Cluster analysis of gene expression data has proved to be a useful tool for identifying co-expressed genes. DNA microarrays are emerged as the leading technology to measure gene expression levels primarily, because of their high throughput. Results from these experiments are usually presented in the form of a data matrix in which rows represent genes and columns represent conditions [12]. Each entry in the matrix is a measure of the expression level of a particular gene under a specific condition. Analysis of these data sets reveals genes of unknown functions and the discovery of functional relationships between genes [18]. The most popular clustering algorithms in microarray gene expression analysis are Hierarchical clustering [11], K-Means clustering [3], and SOM [8]. Of these K-Means clustering is very simple and fast efficient. This is most popular one and it is developed by Mac Queen [6]. The easiness of K-Means clustering algorithm made this algorithm used in several fields. The K-Means algorithm is effective in producing clusters for many practical applications, but the computational complexity of the original K-Means algorithm is very high, especially for large data sets. The K-Means clustering algorithm is a partitioning clustering method that separates the data into K groups. One drawback in the K-Means algorithm is that of a priori fixation of number of clusters [2, 3, 4, 17].

Iterative Self-Organizing Data Analysis Techniques (ISODATA) tries to find the best cluster centres through iterative approach, until some convergence criteria are met. One significant feature of ISODATA over K-Means is that the initial number of clusters may be merged or split, and so the final number of clusters may be different from the number of clusters specified as part of the input. The ISODATA requires number of clusters, and a number of additional user-supplied parameters as inputs. To get better results user need to initialize these parameters with appropriate values by analyzing the input microarray data. In [10] Karteeka Pavan et al proposed AGMFI algorithm to initialize merge factor for ISODATA. This paper studies an initialization of centroids proposed in [17] for microarray data to get the best quality of clusters. A comparative analysis is performed for UCI data sets in-order to get the best clustering algorithm.

This paper is organised as follows. Section 2 presents an overview of Existing works K-Means algorithm, Automatic Generation of Merge Factor for ISODATA (AGMFI) methods. Section 3 describes the Enhanced initialization algorithm. Section 4 describes performance study of the above methods for UCI data sets. Section 5 describes the conclusion and future work.

## 2. Related Work

2.1 K- Means Clustering

The main objective in cluster analysis is to group objects that are similar in one cluster and separate objects that are dissimilar by assigning them to different clusters. One of the most popular clustering methods is K-Means clustering algorithm [3, 9, 12, 17]. It classifies object to a pre-defined number of clusters, which is given by the user (assume *K* clusters). The idea is to choose random cluster centres, one for each cluster. These centres are preferred to be as far as possible from each other. In this algorithm mostly Euclidean distance is used to find distance between data points and centroids [6]. The Euclidean distance between two multi-dimensional data points $X = (x_1, x_2, x_3, ..., x_m)$ and $Y = (y_1, y_2, y_3, ..., y_m)$ is described as follows:

$$D(X,Y) = \sqrt{(x_1 - y_1)^2 + (x_2 - y_2)^2 + \cdots + (x_m - y_m)^2}$$

The K-Means method aims to minimize the sum of squared distances between all points and the cluster centre. This procedure consists of the following steps, as described below.
___________________________________________
**Algorithm 1**: K-Means clustering algorithm [17]
___________________________________________

**Require**: D = {$d_1, d_2, d_3, ..., d_n$} // Set of n data points.
   K - Number of desired clusters
**Ensure**: A set of K clusters.

**Steps:**
1. Arbitrarily choose *k* data points from *D* as initial centroids;
2. **Repeat**
   Assign each point $d_i$ to the cluster which has the closest centroid;
   Calculate the new mean for each cluster;
   **Until** convergence criteria is met.
___________________________________________

Though the K-Means algorithm is simple, it has some drawbacks of quality of the final clustering, since it highly depends on the arbitrary selection of the initial centroids [1].

2.2 Automatic Generation of Merge Factor for ISODATA (AGMFI) Algorithm

The clusters produced in the K-Means clustering are further optimized by ISODATA algorithm. Some of the parameters are fixed by user during the merging and partitioning the clusters. In [10], Automatic Generation of Merge Factor is proposed to initialize merge factor for ISODATA. AGMFI uses different heuristics to determine when to split. Decision of merging is done based upon merge factor which is the function of distances between the clusters. The step by step procedure of AGMFI is given here under.
___________________________________________
**Algorithm 3**: The AGMFI algorithm [10]
___________________________________________

**Require**: D = {$d_1, d_2, d_3, ..., d_n$} // Set of n data points.
   K - Number of desired clusters.
   m- minimum number of samples in a cluster.
   N – maximum number of iterations.
   Θs – a threshold value for spilt_size.
   Θc - a threshold value for merge_size.
**Ensure**: A set of K clusters.

**Steps**:
1. Identify clusters using K-Means algorithms;
2. Find the inter distance in all other cluster to minimum average inter distances clusters point in C;
3. Discard the m and merging operations of cluster ≥ 2*K, If n is even go to step 4 or 5;
4. Distance between two centroids < C, merge the Cluster and update centroid, otherwise repeat up to K/2 times;
5. K ≤ K/2 or n is odd go to step 6 or 7;
6. Find the standard division of all clusters that has exceeds S * standard division of D;
7. Executed N times or no changes occurred in clusters since the last time then stop, otherwise take the centroids of the clusters as new seed points and find the clusters using K-Means and go to step 3.
___________________________________________

The main difference between AGMFI and ISODATA is ISODATA uses heuristic values to merge the clusters, AGMFI generates automatically and the choice of c is not fixed but is to be decided to have better performance. The distance measure used here is the Euclidean distance. To assess the quality of the clusters, we used the silhouette measure proposed by Rousseeuw [14].

## 3. The Enhanced Method

Performance of iterative clustering algorithms which converges to numerous local minima depends highly on initial cluster centers. Generally initial cluster centers are selected randomly. In this section, the cluster centre initialization algorithm is studied to improve the performance of the K-Means algorithm.

**Algorithm 2:** The Enhanced Method [17]

**Require:** $D = \{d_1, d_2, d_3, ..., d_i, ..., d_n\}$ // Set of n data points.
$d_i = \{x_1, x_2, x_3, ..., x_i, ..., x_m\}$ // Set of attributes of one data point.
$k$ // Number of desired clusters.
**Ensure:** A set of $k$ clusters.
**Steps:**
1. In the given data set $D$, if the data points contains the both positive and negative attribute values then go to Step 2, otherwise go to step 4.
2. Find the minimum attribute value in the given data set $D$.
3. For each data point attribute, subtract with the minimum attribute value.
4. For each data point calculate the distance from origin.
5. Sort the distances obtained in step 4. Sort the data points accordance with the distances.
6. Partition the sorted data points into k equal sets.
7. In each set, take the middle point as the initial centroid.
8. Compute the distance between each data point $d_i$ ($1 \le i \le n$) to all the initial centroids $c_j$ ($1 \le j \le k$).
9. **Repeat**
10. For each data point $d_i$, find the closest centroid $c_j$ and assign $d_i$ to cluster $j$.
11. Set ClusterId[$i$]=j. // j:Id of the closest cluster.
12. Set NearestDist[$i$]= $d(d_i, c_j)$.
13. For each cluster $j$ ($1 \le j \le k$), recalculate the centroids.
14. **For** each data point $d_i$,
14.1 Compute its distance from the centroid of the present nearest cluster.
14.2 If this distance is less than or equal to the present nearest distance, the data point stays in the same cluster.
   Else
14.2.1 For every centroid $c_j$ ($1 \le j \le k$) compute the distance $d(d_i, c_j)$.
   End for;
   **Until** the convergence criteria is met.

## 4. Experimental Analysis and Discussion

The following data sets are used to analyse the methods studied in sections 2 and 3.

*Serum data*
This data set is described and used in [10]. It can be downloaded from: http://www.sciencemag.org/feature/data/984559.shl and corresponds to the selection of 517 genes whose expression varies in response to serum concentration inhuman fibroblasts.

*Yeast data*
This data set is downloaded from Gene Expression Omnibus-databases. The Yeast cell cycle dataset contains 2884 genes and 17 conditions. To avoid distortion or biases arising from the presence of missing values in the data matrix we removal all the genes that had any missing value. This step results in a matrix of size 2882 * 17.

*Simulated data*
It is downloaded from http://www.igbmc.ustrasbg.fr/projets/fcm/y3c.txt. The set contains 300 Genes [3]. Above the microarray data set values are all normalized in every gene average values zero and standard deviation equal to 1.

4.1 Comparative Analysis

The K-Means, Enhanced with K-Means and AGMFI are applied on serum data set when number of clusters is taken as 10 and 5 times running to EAGMFI clusters data into 7.

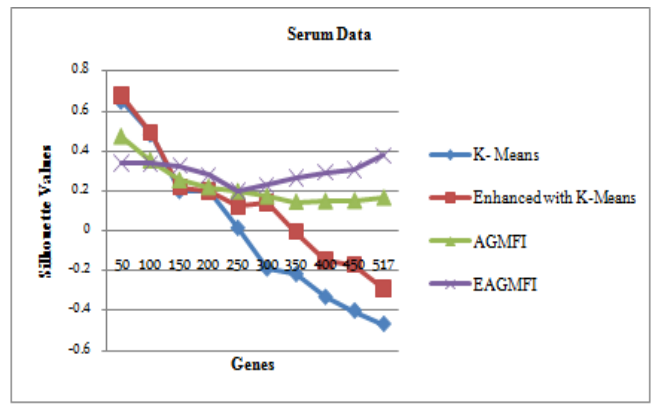

Fig. 1 Performance Comparison chart for serum data.

K-Means, Enhanced with K-Means and AGMFI are applied on Yeast data set when number of clusters initialized to 34 and 5 times running on EAGMFI clusters data into 18 .

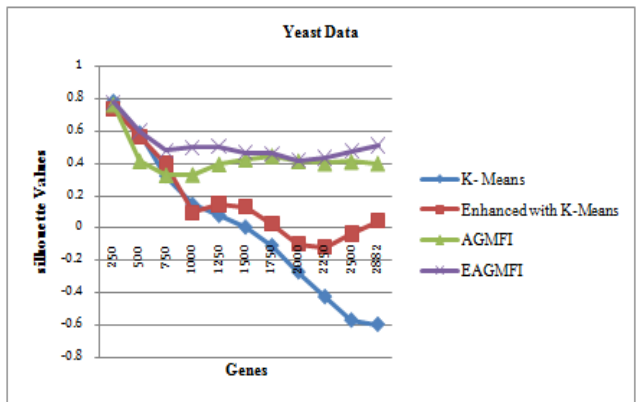

Fig. 2 Performance Comparison chart for Yeast data.

The K-Means, Enhanced with K-Means and AGMFI are applied on simulated data set when number of clusters initialized to 10 and 5 times running to EAGMFI clusters data into 6.

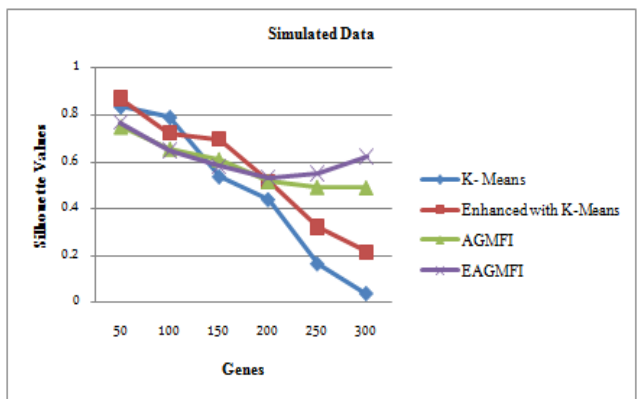

Fig. 3 Performance Comparison chart for simulated data.

Table 1: Comparative Analysis of Clustering Quality of Measurement.

| Data set | Initial No of cluster | Finalized No of cluster | Cluster Quality by K-Means | Cluster Quality by Enhanced with K-Means | Cluster Quality by AGMFI | Cluster Quality by EAGMFI |
|---|---|---|---|---|---|---|
| Serum | 10 | 7 | -0.566 | 12.38 | 22.78 | 29.594 |
| Yeast | 34 | 18 | -0.43 | 17.27 | 43.26 | 51.38 |
| simulated | 10 | 6 | 46.74 | 55.79 | 58.695 | 61.79 |

It is observed from the above analysis that the proposed method is performing well for all the three data cells.

## 5. Conclusion

In this paper AGMFI was studied to improve the quality of clusters. The Enhanced Automatic Generation of Merge Factor for ISODATA (EAGMFI) Clustering Microarray Data based on K-Means and AGMFI clustering algorithms were also studied. One of the demerits of AGMFI is random selection of initial seed point of desired clusters. This was overcome with Enhanced for finding the initial centroids algorithms to avoidance for initial values at random. Therefore, the EAGMFI algorithm not depending upon the any choice of the number of cluster and automatic evaluation initial seed of centroids it produces different better results with Silhouette Coefficients measurement. Both the algorithms were tested with gene expression data.